\documentclass[10pt,twocolumn,letterpaper]{article}

\usepackage[pagenumbers]{cvpr} 

\definecolor{cvprblue}{rgb}{0.21,0.49,0.74}
\usepackage[pagebackref,breaklinks,colorlinks,allcolors=cvprblue]{hyperref}

\usepackage{booktabs}
\usepackage{amsmath}
\usepackage{multirow}
\usepackage{pifont}
\usepackage[table]{xcolor}

\def\confName{CVPR}
\def\confYear{2027}

\title{Inspect Before You Decide: Evidence-Guided Visual Search for Industrial Anomaly Detection}

\author{
Peng Chen$^{1}$ \quad Chao Huang$^{1}$ \quad
Yangyang Liu$^{1}$ \quad Maodong Wu$^{1}$ \quad
Jinyi Xu$^{1}$ \\ Mingbo Yang$^{1}$ \quad
Yunkang Cao$^{2}$ \quad Fangjun Huang$^{1}$ \\
\\
$^{1}$Shenzhen Campus of Sun Yat-sen University \quad
$^{2}$Hunan University
}

\begin{document}
\maketitle
\begin{abstract}
Recent advances in multimodal large language models (MLLMs) have demonstrated remarkable potential for industrial anomaly detection and understanding. However, reliably identifying subtle anomalies remains challenging, as existing approaches often struggle to acquire and verify fine-grained visual evidence beyond global observations. To address this challenge, we propose EVAD, an evidence-guided visual search framework that enables MLLMs to actively seek and examine diagnostically informative visual evidence for reliable anomaly reasoning. Specifically, EVAD begins with a global visual assessment to identify potentially anomalous regions for further inspection. Building upon this observation, we introduce a Monte Carlo Tree Search (MCTS)-guided exploration strategy that progressively acquires visual evidence through two complementary operations: semantic refinement for fine-grained inspection and spatial expansion for contextual verification. The acquired evidence is subsequently evaluated and integrated to support more informed and reliable anomaly decisions. Extensive experiments across multiple industrial anomaly detection tasks demonstrate that EVAD achieves state-of-the-art performance in most evaluated settings, highlighting the effectiveness of active evidence exploration for fine-grained anomaly understanding and reasoning. The code will be publicly available upon publication.
\end{abstract}

\section{Introduction}

\begin{figure}
    \centering
    \includegraphics[width=\linewidth]{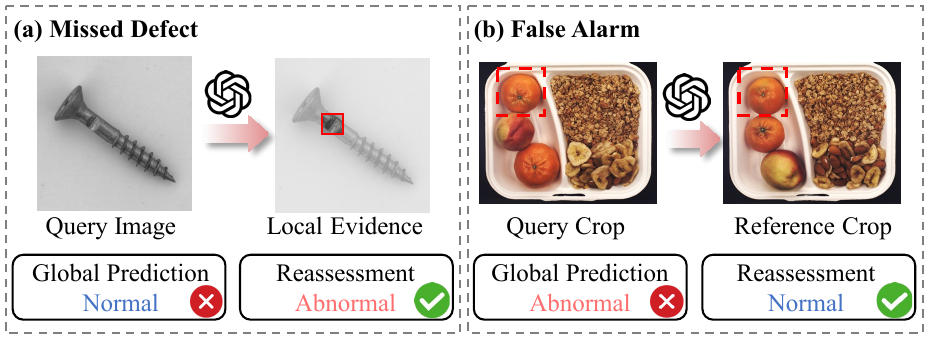}
    \caption{Local visual evidence for reassessing global anomaly predictions. (a) Inspecting the highlighted region on the screw shank reveals a defect missed during global inspection, changing the prediction from normal to abnormal. (b) Comparing the highlighted fruit region with its counterpart in the normal reference supports revising the initial abnormal prediction to normal.}
    \label{fig:intro}
\end{figure}

Industrial anomaly detection is central to automated quality inspection~\cite{bergmann2019mvtec}.
Beyond determining whether a product is defective, an inspection system
often needs to identify the defect type, locate the affected component,
and describe how the product deviates from its expected appearance or
configuration. Conventional methods typically detect such deviations
through normality modeling or reconstruction~\cite{roth2022towards,zavrtanik2021draem}. Vision-language models
(VLMs) extend anomaly detection to zero-shot settings through
transferable visual-semantic representations~\cite{jeong2023winclip,zhou2024anomalyclip,chen2026dyc}, while multimodal large
language models (MLLMs) support a broader range of inspection tasks
through visual question answering~\cite{gu2024anomalygpt,jiang2025mmad}. These tasks remain challenging when
the relevant evidence occupies only a small fraction of the image or
depends on spatial relationships among components~\cite{bergmann2022beyond}.

Local inspection can help correct both missed detections and false
alarms. In Fig.~\ref{fig:intro}(a), the MLLM initially classifies
the screw as normal. A closer view of the shank reveals a defect
overlooked in the full-image assessment. In Fig.~\ref{fig:intro}(b),
the appearance of the fruit prompts an initial alarm. Comparing
the flagged region with its counterpart in a normal reference
supports interpreting the apparent deviation as normal variation.
These cases illustrate an asymmetry in the evidence required to
revise an initial prediction. A single region showing a clear defect
may suffice to overturn a normal prediction, even when the defect
is small. Revising an abnormal prediction requires revisiting the
cue that triggered the alarm and finding evidence that it is
consistent with normal appearance or configuration.

Recent methods help MLLMs acquire fine-grained visual evidence through
hierarchical zooming and expert-guided attention~\cite{wu2024vstar,li2025triad}. ZoomEye~\cite{shen2025zoomeye} searches a
hierarchy of image crops for question-relevant details, and EAGLE~\cite{peng2026eagle} uses
an expert detector to highlight potentially defective regions. In
industrial inspection, the appropriate field of view depends on the
cue under examination. A suspected scratch may require a close-up of
the component surface, whereas a suspected assembly error may require
a wider view that shows how the component is positioned relative to its
neighbors~\cite{bergmann2022beyond}. Under a fixed inspection budget, the search must determine
where to inspect and how much surrounding context to include. Evidence
from earlier observations can guide these choices by indicating which
regions warrant further examination.

Once local observations have been collected, their usefulness depends
on their relevance to the initial judgment and on the information they
add. Overlapping crops can repeat the same ambiguous cue, and a clear
view of an unrelated component offers little basis for changing the
prediction. Effective reassessment therefore requires selecting
complementary observations and interpreting them alongside the full
query image and the normal reference. This allows the model to relate
local findings to the configuration of the product as a whole.

We propose EVAD, a training-free framework that actively acquires local
visual evidence to reassess global predictions for industrial anomaly
understanding. EVAD first makes a global prediction and uses a
pretrained visual grounding model to localize semantic inspection
targets. It then explores candidate regions through budgeted Monte
Carlo tree search~\cite{kocsis2006bandit}. Semantic refinement grounds more specific targets
for closer inspection, while spatial expansion broadens the field of
view to capture surrounding context. For each region, the MLLM summarizes the visible cues in text,
assesses whether the view provides sufficient detail and context,
and separately scores the evidence supporting abnormality and normality. A
validity-weighted reward gives greater weight to abnormal evidence
while allowing strong normal evidence to guide subsequent exploration.
For final reassessment, EVAD filters weak observations and penalizes
spatial overlap during evidence selection. The MLLM reviews the selected
evidence alongside the full query image and the normal reference,
retaining the initial prediction when the evidence is insufficient
to justify a revision. On MMAD~\cite{jiang2025mmad}, EVAD improves the seven-task mean accuracy of Qwen3-VL-8B~\cite{bai2025qwen3vl}
from 74.5\% to 81.4\%, with gains of 13.7 and 11.3 percentage points
in defect classification and defect localization, respectively.
Our contributions are summarized as follows:

\begin{itemize}
    \item We introduce EVAD, a training-free framework for industrial
    anomaly understanding that grounds semantic inspection targets
    and uses local visual evidence to reassess global predictions.

    \item We develop a bidirectional evidence-guided tree search that
    combines semantic refinement and spatial expansion and uses a
    validity-weighted reward to allocate the inspection budget based
    on evidence of abnormality and normality.

    \item We design a diversity-aware reassessment strategy that
    selects local observations based on evidence strength and spatial
    overlap and evaluates whether they warrant revising the initial
    prediction.
\end{itemize}

\section{Related Work}

\subsection{Anomaly Detection}

Industrial anomaly detection has traditionally focused on identifying
and localizing deviations from normal appearance~\cite{cao2024adaclip,gao2026adaptclip}. PatchCore~\cite{roth2022towards} compares
test patch features with a memory bank of normal features. Vision-language
models (VLMs) extend anomaly classification and localization to zero-shot
settings. WinCLIP~\cite{jeong2023winclip} matches multi-scale visual features with compositional
prompts, while AnomalyCLIP~\cite{zhou2024anomalyclip} learns object-agnostic prompts for normal and
abnormal states. Multimodal large language models (MLLMs) extend industrial
inspection to defect description and analysis, which MMAD~\cite{jiang2025mmad} evaluates through
visual question answering. AnomalyGPT~\cite{gu2024anomalygpt} supports interactive defect inspection,
and Triad~\cite{li2025triad} combines expert-guided region tokenization with manufacturing
knowledge. IAD-R1~\cite{li2026iadr1} uses supervised fine-tuning followed by reinforcement
learning, while AD-FM~\cite{liao2026adfm} combines multi-stage reasoning with rewards for
classification and localization. At inference time, ADSeeker~\cite{zhang2026adseeker} retrieves
domain knowledge from visual documents, and Reason-IAD~\cite{chen2026towards} combines knowledge
retrieval with iterative latent reasoning and selective injection of image
patches.

\subsection{Visual Reasoning}

Fine-grained visual reasoning often requires examining details and spatial
relationships that are difficult to resolve from a global view~\cite{liu2025hide,li2026deepscan,li2026bvs}. V*~\cite{wu2024vstar} uses
language-guided search to locate task-relevant regions, while Visual
Sketchpad~\cite{hu2024visual} uses visual annotations and specialized perception tools to
support intermediate reasoning. VisuoThink~\cite{wang2025visuothink} explores alternative
visual-textual reasoning paths through look-ahead tree search, and ZoomEye~\cite{shen2025zoomeye}
searches a hierarchy of image crops for question-relevant details. In
industrial inspection, EAGLE~\cite{peng2026eagle} converts predictions from a PatchCore-based
expert into visual prompts that direct the MLLM to suspected defects.
Local inspection can reveal a defect missed in the global view or show,
through comparison with a normal reference, that a suspicious feature is
consistent with normal variation. EVAD uses evidence supporting either
interpretation to guide subsequent exploration and reassess the initial
prediction.

\begin{figure*}
    \centering
    \includegraphics[width=\linewidth]{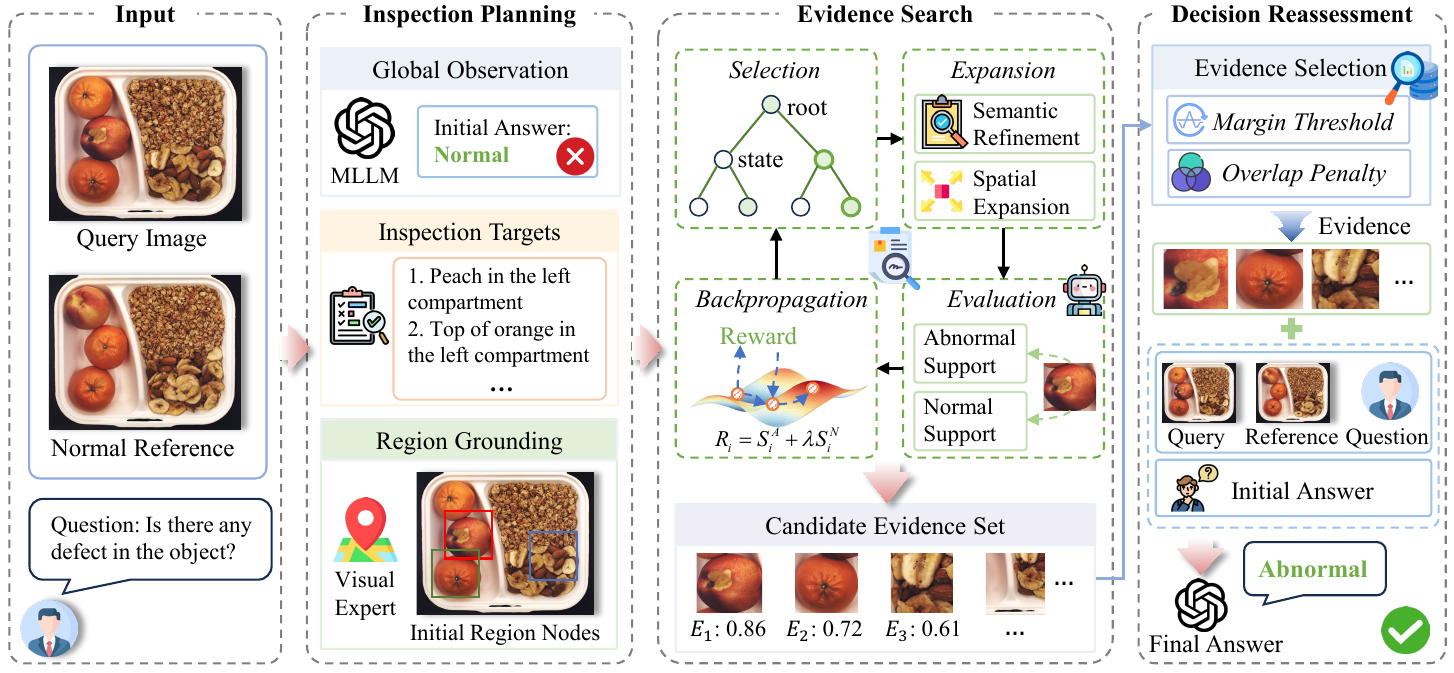}
    \caption{Overview of the proposed EVAD framework. EVAD consists of three key stages: (1) Inspection Planning, which performs global observation and generates spatially grounded semantic inspection targets; (2) Evidence Search, which employs Monte Carlo Tree Search (MCTS) with bidirectional evidence evaluation and asymmetric rewards to explore informative visual regions; and (3) Decision Reassessment, which selects diverse and informative visual evidence to reassess the initial prediction.}
    \label{fig:method}
\end{figure*}

\section{Methodology}

\subsection{Problem Definition}

We consider industrial anomaly understanding in a multimodal question-answering setting, covering anomaly detection, localization, and reasoning. Given a query image $I_q \in \mathbb{R}^{H \times W \times 3}$, a normal reference image $I_r$ from the same object category, an inspection question $Q$, and a set of candidate answers $\mathcal{A} = \{a_1, a_2, \ldots, a_C\}$, the goal is to predict the correct answer $\hat{a} \in \mathcal{A}$. Formally, the prediction is formulated as:
\begin{equation}
    \hat{a} = \mathcal{F}(I_q, I_r, Q, \mathcal{A}),
\end{equation}
where $\mathcal{F}$ denotes the multimodal prediction function for industrial anomaly understanding.

\subsection{Overview of EVAD}

To improve fine-grained anomaly understanding, we propose EVAD, an evidence-guided visual search framework that enhances initial predictions through active acquisition and verification of informative visual evidence. As illustrated in Fig.~\ref{fig:method}, EVAD comprises three key stages: (1) Semantics-Guided Inspection Planning, (2) Bidirectional Evidence-Guided Tree Search, and (3) Diversity-Aware Decision Reassessment.

Specifically, EVAD begins with a global inspection to establish an initial prediction and identify potential inspection targets, which are subsequently grounded into candidate regions for visual exploration. A bidirectional evidence-guided tree search then progressively examines these regions through semantic refinement and spatial expansion, evaluating both normal and abnormal evidence to guide exploration toward diagnostically informative regions. Finally, a diversity-aware selection strategy identifies complementary evidence while suppressing redundant observations. The selected evidence is integrated with the global visual context and initial prediction to reassess the initial decision and produce the final answer.

\subsection{Semantics-Guided Inspection Planning}

Fine-grained industrial anomalies are often subtle and spatially localized, making them difficult to identify through global inspection alone. To address this challenge, we introduce a semantics-guided inspection planning strategy that translates global visual understanding into targeted regional inspection.

Given a query image $I_q$, a normal reference image $I_r$, and
an inspection prompt $Q$, an MLLM $f_{\theta}$ performs
a global inspection and produces an initial prediction:
\begin{equation}
    a_0 = f_{\theta}(I_q, I_r, Q, \mathcal{A}),
\end{equation}
where $\mathcal{A}$ denotes the candidate answer set. Rather than directly relying on this prediction, we further guide the MLLM to generate a set of semantic inspection targets:
\begin{equation}
    \mathcal{T} = \{t_1, t_2, \ldots, t_M\},
\end{equation}
where each $t_i$ specifies an object component or a potentially informative region requiring closer examination. 

To associate these semantic targets with specific image regions, we employ a visual grounding expert $g_{\phi}$ to identify the corresponding regions. Given an inspection target $t_i$, the grounding process is formulated as:
\begin{equation}
    (b_i, m_i) = g_{\phi}(I_q, t_i),\quad i = 1, \ldots, M,
\end{equation}
where $b_i$ and $m_i$ denote the predicted bounding box and segmentation mask, respectively. The grounded regions constitute an initial candidate set $\mathcal{R}_0 = \{r_i\}_{i=1}^{M}$, where $r_i = (t_i, b_i, m_i)$. These regions initialize the subsequent tree-based search, directing the exploration toward semantically relevant visual evidence.

\subsection{Bidirectional Evidence-Guided Tree Search}

Although semantics-guided inspection planning identifies candidate regions for further examination, these regions may fail to capture subtle defects or sufficient contextual information. To address this limitation, we use Monte Carlo Tree Search (MCTS) to explore additional regions under a fixed budget, guided by local evidence supporting either abnormality or normality.

\noindent\textbf{Search Space and Actions.}
Given the initial candidate set $\mathcal{R}_0$, we construct a search tree rooted at the global observation, with the initial regions forming its child nodes. To facilitate progressive visual exploration, we define two complementary actions:
\begin{equation}
    \mathcal{U} = \{u_{\mathrm{sem}}, u_{\mathrm{spa}}\},
\end{equation}
where $u_{\mathrm{sem}}$ and $u_{\mathrm{spa}}$ denote semantic refinement and spatial expansion, respectively. Semantic refinement generates a more specific inspection target and leverages the visual grounding expert to localize the corresponding region. In contrast, spatial expansion enlarges the current bounding box to incorporate broader visual context. Together, these actions enable the search to capture fine-grained details and contextual relationships that may be overlooked during initial inspection.

\noindent\textbf{Bidirectional Evidence Evaluation.}
For each candidate region $r_i=(t_i,b_i,m_i)$, we extract a local
view $I_i^{\mathrm{loc}}=\operatorname{Crop}(I_q,b_i)$ from the
query image. Given the inspection target $t_i$ and the
question-specific prompt $Q$, the MLLM compares the target in
the local view with the corresponding component or location in
the normal reference image $I_r$. In a single evaluation, the
model generates a textual observation $o_i$ and three numerical
ratings:
\begin{equation}
    (o_i,V_i,A_i,N_i)
    =
    f_{\theta}^{\mathrm{eval}}
    \bigl(I_i^{\mathrm{loc}}, I_r, t_i, Q\bigr),
     V_i,A_i,N_i\in[0,1],
\end{equation}
where $f_{\theta}^{\mathrm{eval}}$ denotes the same MLLM
conditioned on a region evaluation prompt. The observation $o_i$
summarizes the visible cues for $t_i$. The validity score $V_i$
rates whether the local view provides sufficient detail and
context to assess the target, from insufficient visual information
($0$) to a clearly assessable view ($1$). Abnormal support $A_i$
rates the visible evidence of a defect or structural inconsistency,
whereas normal support $N_i$ rates the visible evidence that the
target's appearance or configuration is consistent with the
reference. For each support score, $0$ indicates no visible support
and $1$ indicates strong support. The model assigns the two support
scores separately, allowing both to be low when the visual evidence
is inconclusive.

We weight the difference between the support scores by the
validity of the observation and derive two directional evidence
scores:
\begin{equation}
    \Delta_i = V_i(A_i-N_i), \quad
    \bigl(S_i^{A},S_i^{N}\bigr)
    = \bigl([\Delta_i]_+,[-\Delta_i]_+\bigr),
\end{equation}
where $[x]_+=\max(0,x)$. A positive margin indicates that the
observation favors abnormality, and a negative margin indicates
that it favors normality. Its magnitude decreases as the two
support scores become closer or the validity of the observation
decreases. We store each evaluated region as an evidence record
$e_i=(r_i,I_i^{\mathrm{loc}},o_i,\Delta_i)$ for subsequent evidence
selection and decision reassessment.

To prioritize the inspection of potential defects, we define
the search reward as:
\begin{equation}
    R_i = S_i^{A} + \lambda S_i^{N},
    \qquad 0<\lambda<1,
\end{equation}
where $\lambda$ controls the contribution of normal evidence.
For margins of equal magnitude, the reward is larger in the
abnormal direction. Regions with strong evidence of normality
can still contribute to the search.

\noindent\textbf{Tree Search and Reward Backpropagation.}
We initialize the tree with a root representing the global
observation and attach each successfully grounded region in
$\mathcal{R}_0$ as a child of the root. We evaluate each initial
region once and backpropagate its reward to the root. Thus, each
initial child $i$ starts with visit count $n_i=1$ and mean reward
$\bar{R}_i=R_i$. We also add its evidence record to the candidate
set $\mathcal{E}$.

Each search iteration starts at the root and descends the tree
until it reaches a node with an untried valid action. When the
current node has no such action, we select the child with the
highest upper confidence bound (UCB) among those whose subtrees
contain an expandable node:
\begin{equation}
    \operatorname{UCB}(i)
    = \bar{R}_i
    + c \sqrt{\frac{\ln(n_p+1)}{n_i+\epsilon}},
\end{equation}
where $n_i$ and $n_p$ are the visit counts of node $i$ and its
parent $p$, respectively, $\bar{R}_i$ is the mean reward
backpropagated through node $i$, $c$ controls exploration, and
$\epsilon$ is a small positive constant.

At the selected node, we attempt semantic refinement first if it is
available, followed by spatial expansion. Each action is attempted
at most once per node. If an action fails to produce a valid region,
we mark it as tried and continue the search. Otherwise, we evaluate
the new child $j$, add its evidence record to $\mathcal{E}$, and
backpropagate $R_j$ along the path from $j$ to the root, updating the
visit count and mean reward at each node on the path.

The search terminates after evaluating $B$ additional regions or
when no expandable node remains. The resulting candidate set
$\mathcal{E}$ contains evidence records for all evaluated regions,
including the initial grounded regions and those generated through
semantic refinement or spatial expansion.

\subsection{Diversity-Aware Decision Reassessment}

Although bidirectional evidence-guided tree search identifies informative visual regions, the resulting evidence may contain redundant or ambiguous observations that compromise subsequent reasoning. To address this issue, we introduce a diversity-aware decision reassessment strategy that selects complementary visual evidence to support reliable prediction refinement.

\noindent\textbf{Diversity-Aware Evidence Selection.}
For each candidate $e_i \in \mathcal{E}$ associated with region
$r_i$, we measure its evidence strength using the magnitude of
the signed evidence margin:
\begin{equation}
    E_i = |\Delta_i| = \max\bigl(S_i^{A}, S_i^{N}\bigr).
\end{equation}
The search reward favors abnormal evidence of equal strength
to prioritize the inspection of potential defects. At the
selection stage, $E_i$ gives equal weight to both directions,
so that either can inform reassessment. We retain candidates
whose evidence strength meets a threshold $\tau$:
\begin{equation}
    \mathcal{C} = \{i \mid e_i \in \mathcal{E},\ E_i \geq \tau\}.
\end{equation}

Starting from $\mathcal{S}_0 = \varnothing$, we greedily select
$L = \min(K, |\mathcal{C}|)$ candidates, where $K$ is the maximum
number of retained observations. At step $k$, we select
\begin{equation}
\begin{aligned}
    i_k &\in \operatorname*{arg\,max}_{i \in
        \mathcal{C} \setminus \mathcal{S}_{k-1}}
        \left[
            E_i - \gamma \max_{j \in \mathcal{S}_{k-1}}
            \operatorname{IoU}(b_i,b_j)
        \right], \\
    \mathcal{S}_k &= \mathcal{S}_{k-1} \cup \{i_k\},
\end{aligned}
\end{equation}
where $b_i$ is the bounding box of region $r_i$ and $\gamma$
controls the overlap penalty. The penalty is zero when
$\mathcal{S}_{k-1}$ is empty. At subsequent steps, each candidate
is penalized according to its largest overlap with any selected
region. The resulting evidence set is
$\mathcal{E}^{*} = \{e_i \mid i \in \mathcal{S}_L\}$.

\noindent\textbf{Decision Reassessment.}
We reassess the initial answer $a_0$ by interpreting the selected
local evidence $\mathcal{E}^{*}$ in the context of the query image
$I_q$ and the normal reference image $I_r$.

For anomaly discrimination, we instruct the MLLM to revise an
initially normal prediction only when the selected evidence reveals
a defect or a structural inconsistency. To revise an initially
abnormal prediction, the MLLM re-examines the query image for
apparent defect cues and checks whether the selected evidence
directly addresses those cues and supports a normal interpretation.
For other inspection questions, the prompt allows a revision only
when the evidence favors an alternative answer over $a_0$.
The MLLM is instructed to retain $a_0$ when the evidence is
insufficient to justify a revision. The final answer is given by:
\begin{equation}
    \hat{a} =
    \begin{cases}
        a_0, & \mathcal{E}^{*} = \varnothing, \\
        f_{\theta}\bigl(I_q, I_r, Q, \mathcal{A}, a_0,
        \mathcal{E}^{*}\bigr), & \text{otherwise}.
    \end{cases}
\end{equation}

The full query image situates each local observation within the
object's overall structure, and the normal reference provides
a basis for distinguishing subtle defects from normal variations
in appearance or configuration.

\section{Experiments}

\begin{table*}[t]
\centering
\caption{Comparison with representative MLLMs and anomaly detection methods on multiple evaluation tasks. The best results are highlighted in bold, and all metrics are reported as accuracy (\%).}
\label{tab:main_results}
\setlength{\tabcolsep}{5pt}
\renewcommand{\arraystretch}{1.05}
\resizebox{\textwidth}{!}{%
\begin{tabular}{lcccccccc}
\toprule
\multirow{2}{*}{Method}
& \multicolumn{1}{c}{Anomaly}
& \multicolumn{4}{c}{Defect}
& \multicolumn{2}{c}{Object}
& \multirow{2}{*}{Average} \\
\cmidrule(lr){2-2}
\cmidrule(lr){3-6}
\cmidrule(lr){7-8}
& Discrimination & Classification & Localization
& Description & Analysis
& Classification & Analysis & \\
\midrule
\rowcolor{gray!12}
\multicolumn{9}{l}{\textbf{\textit{Human evaluation}}} \\
\midrule
Human (Expert)
& 95.2 & 75.0 & 92.3 & 83.3
& 94.2 & 86.1 & 80.4 & 86.7 \\
Human (Ordinary)
& 86.9 & 66.3 & 85.6 & 71.3
& 81.5 & 89.6 & 69.7 & 78.7 \\

\midrule
\rowcolor{gray!12}
\multicolumn{9}{l}{\textbf{\textit{Proprietary Models}}} \\
\midrule

Claude-3.5-Sonnet
& 60.1 & 60.1 & 48.8 & 67.1
& 79.1 & 85.2 & 79.8 & 68.6 \\

Gemini-2.5-Flash
& 83.4 & 69.9 & 63.3 & 76.4
& 81.6 & 94.0 & 82.0 & 78.7 \\

Gemini-2.5-Pro
& 83.1 & 73.9 & 67.2 & 80.0
& 86.3 & 94.9 & 83.1 & 81.2 \\

GPT-4o
& 68.6 & 65.8 & 55.6 & 73.2
& 83.4 & 95.0 & 82.8 & 74.9 \\

GPT-5.5
& 70.0 & 64.9 & 62.3 & 76.0
& 80.1 & 95.2 & 82.1 & 75.8 \\

\midrule
\rowcolor{gray!12}
\multicolumn{9}{l}{\textbf{\textit{General-Purpose Open-Source Models}}} \\
\midrule

Qwen3-VL-4B
& 74.1 & 59.7 & 63.6 & 70.1
& 80.0 & 92.4 & 83.4 & 74.8 \\

LLaVA-OneVision-7B
& 56.1 & 55.6 & 45.8 & 65.8
& 75.5 & 91.3 & 81.5 & 67.4 \\

Qwen2.5-VL-7B
& 71.8 & 55.2 & 69.0 & 64.5
& 78.9 & 92.6 & 84.7 & 73.8 \\

MiMo-VL-7B-RL
& 68.7 & 56.0 & 57.3 & 70.1
& 77.1 & 88.5 & 77.8 & 70.8 \\

InternVL3-8B
& 66.3 & 56.4 & 51.2 & 67.7
& 80.3 & 87.8 & 83.7 & 70.5 \\

Qwen3-VL-8B
& 73.1 & 62.3 & 60.3 & 69.1
& 80.1 & 92.5 & 84.0 & 74.5 \\

Qwen3.5-9B
& 68.7 & 62.4 & 59.3 & 73.8
& 81.7 & 91.5 & 85.5 & 74.7 \\

GLM-4.1V-9B-Thinking
& 61.6 & 60.6 & 62.2 & 74.0
& 82.4 & 93.7 & 83.1 & 73.9 \\

Kimi-VL-16B
& 66.6 & 71.7 & 61.9 & 79.8
& 82.9 & 92.2 & \underline{86.5} & 77.4 \\

Qwen2.5-VL-32B
& 70.9 & 58.3 & 62.4 & 65.4
& 83.2 & 76.9 & 84.9 & 71.7 \\

LLaVA-NeXT-34B
& 57.9 & 48.8 & 52.9 & 71.3
& 80.3 & 81.1 & 77.8 & 67.2 \\

InternVL2-76B
& 68.3 & 54.2 & 56.7 & 66.3
& 80.5 & 86.4 & 82.9 & 70.8 \\

\midrule
\rowcolor{gray!12}
\multicolumn{9}{l}{\textbf{\textit{Anomaly-Specialized Training-Based Models}}} \\
\midrule

AnomalyGPT
& 65.6 & 27.5 & 28.0 & 36.9
& 32.1 & 29.8 & 35.8 & 36.5 \\

EIAD
& 60.5 & 50.7 & 55.6 & 67.8
& 76.9 & 91.8 & 82.7 & 69.4 \\

AnomalyR1
& 60.6 & 63.6 & \underline{70.1} & \textbf{80.5}
& \underline{85.3} & 92.5 & 86.2 & 77.0 \\

ADSeeker
& 68.6 & 53.8 & 55.6 & 60.5
& 77.4 & 89.9 & 83.5 & 69.9 \\

\midrule
\rowcolor{gray!12}
\multicolumn{9}{l}{\textbf{\textit{Optimization-based Models}}} \\
\midrule

EAGLE
& 71.6 & 63.7 & 70.0 & \underline{79.9}
& \underline{85.3} & 93.8 & \textbf{87.0} & 78.7 \\

Reason-IAD
& \underline{75.2} & \underline{75.1} & 62.8 & 76.4
& \textbf{85.6} & \underline{96.3} & 84.6 & \underline{79.4} \\

\midrule
\rowcolor{cyan!12}
\textbf{EVAD (Ours, Training-free)}
& \textbf{78.9} & \textbf{76.0} & \textbf{71.6}
& 78.2 & 83.0
& \textbf{96.5} & 85.4 & \textbf{81.4} \\

\rowcolor{cyan!12}
\textit{Gap to best trained}
& $10.3\uparrow$ & $12.4\uparrow$
& $1.5\uparrow$ & $2.3\downarrow$
& $2.3\downarrow$ & $4.0\uparrow$
& $0.8\downarrow$ & $4.4\uparrow$ \\

\bottomrule
\end{tabular}%
}
\end{table*}

\subsection{Experiment Setup}

\noindent
\textbf{Benchmarks.}
We evaluate EVAD on MMAD~\cite{jiang2025mmad}, a benchmark for
industrial anomaly understanding that draws images from
MVTec-AD~\cite{bergmann2019mvtec}, VisA~\cite{zou2022spot},
MVTec-LOCO~\cite{bergmann2022beyond}, and
GoodsAD~\cite{zhang2024goodsad}. MMAD contains 8,366 images and
39,672 multiple-choice questions. Its seven tasks are anomaly
discrimination, defect classification, defect localization,
defect description, defect analysis, object classification, and
object analysis. We adopt the one-shot setting, in which each query
image is paired with one normal reference image from the same object
category. We report the averaged accuracy (\%) across all seven tasks.

\noindent
\textbf{Compared Methods.}
We evaluate EVAD on MMAD against three groups of baselines:
proprietary MLLMs, general-purpose open-source MLLMs, and anomaly-specific methods.
The proprietary baselines include Claude~\cite{anthropic2024claude35sonnet},
Gemini~\cite{comanici2025gemini25}, GPT-4o~\cite{openai2024gpt4o},
and GPT-5.5~\cite{openai2026gpt55}. The general-purpose open-source baselines include models from the
Qwen~\cite{bai2025qwen3vl}, LLaVA~\cite{li2025llavaonevision},
InternVL~\cite{zhu2025internvl3}, GLM~\cite{glmteam2025glmv},
and Kimi-VL~\cite{kimiteam2025kimivl} families, along with
MiMo-VL-7B-RL~\cite{xiaomi2025mimovl}.
The anomaly-specific baselines include AnomalyGPT~\cite{gu2024anomalygpt},
EIAD~\cite{zhang2025eiad}, AnomalyR1~\cite{chao2025anomalyr1},
ADSeeker~\cite{zhang2026adseeker}, EAGLE~\cite{peng2026eagle},
and Reason-IAD~\cite{chen2026towards}.
As reference points, we include the expert and non-expert human
scores reported in the original MMAD benchmark.

\noindent
\textbf{Implementation Details.}
We use Qwen3-VL-8B~\cite{bai2025qwen3vl} as the MLLM backbone
and pretrained SAM3~\cite{carion2026sam3} as the visual grounding
expert. Both models remain frozen throughout evaluation.
The MLLM proposes up to three semantic inspection targets to
initialize the search. We limit the tree depth to three and
evaluate up to 16 additional regions. Each spatial expansion
scales the width and height of the current bounding box by a
factor of 1.35. We set the normal evidence weight in the search
reward to $\lambda=0.5$. For evidence selection, we use a threshold
of $\tau=0.4$ and an overlap penalty coefficient of $\gamma=0.3$,
retaining at most $K=3$ regions for final reassessment. Further
details are provided in the supplementary material.

\subsection{Main Results}

Table~\ref{tab:main_results} presents a comprehensive comparison of EVAD with representative MLLMs and anomaly-specific methods across seven industrial anomaly understanding tasks. Overall, EVAD achieves the highest average accuracy of 81.4\% among all evaluated models, surpassing the strongest proprietary model, Gemini-2.5-Pro, by 0.2 percentage points, and outperforming Reason-IAD and EAGLE by 2.0 and 2.7 percentage points, respectively. Notably, EVAD achieves the best performance in defect classification (76.0\%), defect localization (71.6\%), and object classification (96.5\%), highlighting its effectiveness in fine-grained anomaly understanding. Compared with the strongest training-based baselines, EVAD yields improvements of 10.3, 12.4, and 1.5 percentage points in anomaly discrimination, defect classification, and defect localization, respectively. 
Relative to the frozen Qwen3-VL-8B baseline, EVAD improves all seven
task scores and raises average accuracy from 74.5\% to 81.4\%. Anomaly discrimination improves from 73.1\% to 78.9\%.
The largest gains are in defect classification, localization, and
description, at 13.7, 11.3, and 9.1 percentage points, respectively.
This pattern suggests that local inspection helps the model distinguish
defect types, locate defects, and characterize their appearance.
Additional results are provided in the supplementary material.

\subsection{Ablation Studies}

\textbf{Component Ablation.}
Table~\ref{tab:evad_ablation} evaluates the contributions of tree expansion and guided evidence selection on MMAD. Reassessing the global prediction using only the initial SAM3 regions improves the mean score across the three tasks from 65.2\% to 71.8\%. Enabling tree expansion while keeping the selection and reassessment procedures fixed further raises the mean score to 75.5\%, with gains of 5.3 and 3.7 percentage points in defect classification and localization, respectively. To isolate the effect of guided evidence selection, we uniformly sample regions from the same search candidate pool, match the number of retained regions for each question, and use the same reassessment procedure. Guided selection improves the mean score from 74.5\% to 75.5\%, indicating that it provides more useful evidence for reassessment at the same region count.

\begin{table}[t]
\centering
\caption{Ablation of EVAD on MMAD.
AD denotes accuracy for anomaly discrimination.
DC and DL denote accuracy for defect classification and localization,
respectively. Avg.\ denotes the arithmetic mean of AD, DC, and DL.
All scores are reported in \%.}
\label{tab:evad_ablation}
\setlength{\tabcolsep}{5pt}
\begin{tabular}{@{}lrrrr@{}}
\toprule
Variant & AD & DC & DL & Avg. \\
\midrule
Global prediction     & 73.1 & 62.3 & 60.3 & 65.2 \\
w/o tree expansion
    & 76.8 & 70.7 & 67.9 & 71.8 \\
w/o guided evidence selection
    & 77.8 & 74.9 & 70.7 & 74.5 \\
\textbf{Full EVAD}             & 78.9 & 76.0 & 71.6 & 75.5 \\
\bottomrule
\end{tabular}
\end{table}

\begin{table}[t]
\centering
\caption{Comparison of tree search strategies for local evidence
acquisition on MMAD. EVAD uses upper confidence bound (UCB) to
select branches, while greedy search ($c=0$) disables the exploration
bonus. AD, DC, and DL denote anomaly discrimination, defect
classification, and defect localization, respectively. }
\label{tab:evad_search_strategies}
\setlength{\tabcolsep}{9pt}
\begin{tabular}{@{}lrrrr@{}}
\toprule
Search strategy & AD & DC & DL & Avg. \\
\midrule
Random search & 76.9 & 73.1 & 68.7 & 72.9 \\
Breadth-first search  & 77.2 & 73.8 & 69.3 & 73.4 \\
Greedy search ($c=0$) & 77.7 & 74.6 & 70.1 & 74.1 \\
\textbf{EVAD} & 78.9 & 76.0 & 71.6 & 75.5 \\
\bottomrule
\end{tabular}
\end{table}

\noindent\textbf{Comparison of Search Strategies.}
Table~\ref{tab:evad_search_strategies} compares four tree search
strategies for local evidence acquisition on MMAD. EVAD's UCB-based
search achieves the highest score on all three tasks, with a mean
of 75.5\%, exceeding random, breadth-first, and greedy search by
2.6, 2.1, and 1.4 percentage points, respectively. The greedy variant
uses the same evidence reward and sets the exploration coefficient
to zero ($c=0$), isolating the effect of the UCB exploration bonus.
Enabling this bonus improves AD and DC by 1.2 and 1.4 points,
respectively, with the largest gain on defect localization, from
70.1\% to 71.6\%. This result suggests
that exploring less-visited branches can help uncover spatial cues
that support decision reassessment.

\begin{table}[t]
\centering
\caption{Comparison of EVAD and its backbone on anomaly discrimination
and defect classification across the four MMAD datasets. For each task, Average is the arithmetic mean across the four datasets. Scores are reported in \%.}
\label{tab:evad_per_dataset}
\setlength{\tabcolsep}{5pt}
\begin{tabular}{@{}llrrr@{}}
\toprule
Task & Dataset & Base & EVAD & Gain \\
\midrule
\multirow{5}{*}{\shortstack[l]{Anomaly\\Discrimination}}
    & MVTec-AD   & 86.9 & 91.0 & \textcolor{green!50!black}{+4.1}  \\
    & VisA       & 82.4 & 80.8 & $-1.6$                           \\
    & MVTec-LOCO & 62.7 & 69.5 & \textcolor{green!50!black}{+6.8}  \\
    & GoodsAD    & 60.3 & 74.1 & \textcolor{green!50!black}{+13.8} \\
\cmidrule(lr){2-5}
    & Average    & 73.1 & 78.9 & \textcolor{green!50!black}{\textbf{+5.8}} \\
\midrule
\multirow{5}{*}{\shortstack[l]{Defect\\Classification}}
    & MVTec-AD   & 75.4 & 86.0 & \textcolor{green!50!black}{+10.6} \\
    & VisA       & 58.4 & 73.1 & \textcolor{green!50!black}{+14.7} \\
    & MVTec-LOCO & 51.1 & 69.0 & \textcolor{green!50!black}{+17.9} \\
    & GoodsAD    & 64.4 & 75.8 & \textcolor{green!50!black}{+11.4} \\
\cmidrule(lr){2-5}
    & Average    & 62.3 & 76.0 & \textcolor{green!50!black}{\textbf{+13.7}} \\
\bottomrule
\end{tabular}
\end{table}
\noindent\textbf{Per-Dataset Performance.}
Table~\ref{tab:evad_per_dataset} compares EVAD with the full-image
Qwen3-VL-8B baseline on the four MMAD subsets. Averaged across
subsets, EVAD improves anomaly discrimination from 73.1\% to 78.9\%
and defect classification from 62.3\% to 76.0\%. Defect classification
improves on all four subsets, with gains ranging from 10.6 to
17.9 percentage points and the largest gain on MVTec-LOCO.
Anomaly discrimination improves on three subsets, with the largest
increase on GoodsAD, from 60.3\% to 74.1\%. MVTec-LOCO also gains
6.8 percentage points on this task. On VisA, the two tasks respond
differently to local inspection, with a 14.7-point gain in defect
classification and a 1.6-point decline in anomaly discrimination.

\begin{table}[t]
\centering
\caption{Ablation of search actions on the MVTec-AD and VisA subsets. Semantic and Spatial denote semantic refinement and spatial
expansion, respectively. AD and DC denote anomaly discrimination and
defect classification, respectively.}
\label{tab:evad_action_ablation}
\setlength{\tabcolsep}{10pt}
\begin{tabular}{@{}lccrr@{}}
\toprule
Dataset & Semantic & Spatial & AD & DC \\
\midrule
\multirow{3}{*}{MVTec-AD}
    & \ding{51} & \textcolor{black!40}{\ding{55}} & 89.2 & 84.6 \\
    & \textcolor{black!40}{\ding{55}} & \ding{51} & 90.1 & 82.9 \\
    & \ding{51} & \ding{51}                      & 91.0 & 86.0 \\
\midrule
\multirow{3}{*}{VisA}
    & \ding{51} & \textcolor{black!40}{\ding{55}} & 81.2 & 71.8 \\
    & \textcolor{black!40}{\ding{55}} & \ding{51} & 80.1 & 70.2 \\
    & \ding{51} & \ding{51}                      & 80.8 & 73.1 \\
\bottomrule
\end{tabular}
\end{table}
\noindent\textbf{Effect of Search Actions.} Table~\ref{tab:evad_action_ablation} evaluates semantic refinement and spatial expansion on the MVTec-AD and VisA subsets of MMAD. Combining the two actions yields the highest defect classification accuracy on both subsets, outperforming the best single-action configuration by 1.4 percentage points on MVTec-AD and 1.3 points on VisA. For anomaly discrimination, the combined configuration achieves the highest score on MVTec-AD (91.0\%). On VisA, adding spatial expansion to semantic refinement raises defect classification accuracy from 71.8\% to 73.1\%, while anomaly discrimination decreases from 81.2\% to 80.8\%. The classification gains on both subsets suggest that broader spatial context complements the fine-grained evidence obtained through semantic refinement.

\begin{figure}
    \centering
    \includegraphics[width=\linewidth]{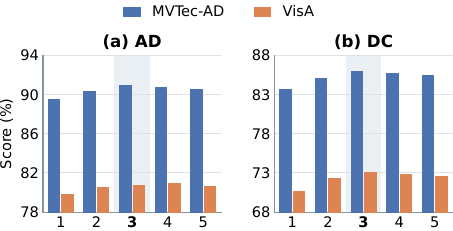}
    \caption{Sensitivity to the maximum number of evidence regions $K$ used for decision reassessment on the MVTec-AD and VisA subsets of MMAD. Panels (a) and (b) show anomaly discrimination (AD) and defect classification (DC), respectively.}
    \label{fig:evad_evidence_topk}
\end{figure}

\begin{figure}
    \centering
    \includegraphics[width=0.9\linewidth]{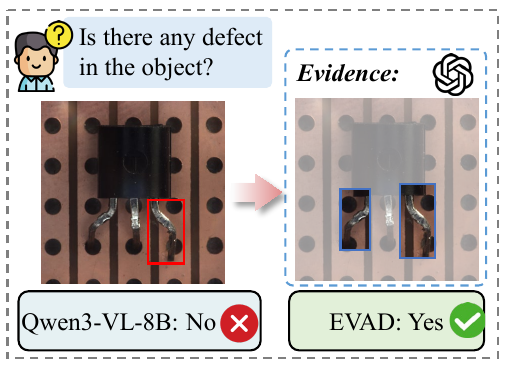}
    \caption{Qualitative example of decision reassessment on the MVTec-AD subset of MMAD. Qwen3-VL-8B initially predicts that the transistor is normal. EVAD examines local evidence around the leads and correctly revises the prediction to anomalous. The red box marks a region inspected by EVAD.}
    \label{fig:case}
\end{figure}

\noindent\textbf{Sensitivity to the Number of Evidence Regions.}
Figure~\ref{fig:evad_evidence_topk} shows the effect of varying the maximum number of evidence regions $K$ used for decision reassessment, with the search candidate pool and selection criteria held fixed. Increasing $K$ from 1 to 3 improves both metrics on MVTec-AD and VisA, with defect classification gains of 2.3 and 2.4 percentage points, respectively. Increasing $K$ beyond 3 changes all scores by at most 0.5 percentage points relative to $K=3$. On VisA, anomaly discrimination peaks at $K=4$ (81.0\%), just 0.2 percentage points above the score at $K=3$. Since defect classification peaks at $K=3$ on both datasets and larger values yield only marginal changes, we use $K=3$ as the default.

\noindent\textbf{Qualitative Analysis.}
Figure~\ref{fig:case} illustrates how EVAD corrects a missed detection
on the MVTec-AD subset of MMAD. In the initial full-image assessment,
Qwen3-VL-8B classifies the transistor as defect-free. The deformation
is confined to the right lead, which occupies only a small portion
of the query image. EVAD examines the highlighted crops around the
base of the transistor, where the shape of the affected lead is more
clearly visible alongside the neighboring leads. During reassessment,
the model relates this local deformation to its position in the full
query image and uses it as evidence of a defect. EVAD consequently
revises the answer from ``No'' to ``Yes,'' in agreement with the
ground-truth label.

\begin{table}[t]
\centering
\caption{Effect of bidirectional evidence on anomaly
discrimination (AD) on the MVTec-AD.
Both reassessment variants use the same initial predictions
and candidate pool. Fix and Error count flips that correct
initial errors and overturn initially correct predictions,
respectively. N and A denote normal and abnormal.}
\label{tab:evad_bidirectional_evidence}
\setlength{\tabcolsep}{6pt}
\begin{tabular}{@{}lccccc@{}}
\toprule
\multirow{2}{*}{Variant} &
\multirow{2}{*}{AD (\%)} &
\multicolumn{2}{c}{$\mathrm{N}\!\to\!\mathrm{A}$} &
\multicolumn{2}{c}{$\mathrm{A}\!\to\!\mathrm{N}$} \\
\cmidrule(lr){3-4}\cmidrule(l){5-6}
& & Fix & Error & Fix & Error \\
\midrule
Global prediction & 86.9 & -- & -- & -- & -- \\
Abnormal only     & 90.3 & 37 & 9  & 8  & 2  \\
Bidirectional     & 91.0 & 38 & 9  & 17 & 5  \\
\bottomrule
\end{tabular}
\end{table}
\noindent\textbf{Effect of Bidirectional Evidence.}
Table~\ref{tab:evad_bidirectional_evidence} compares reassessment
with abnormal evidence only and reassessment with both normal
and abnormal evidence on the MVTec-AD subset of MMAD.
Both variants share the same initial predictions and search
candidate pool. Reassessment with abnormal evidence alone
improves AD from 86.9\% to 90.3\%, and allowing normal evidence
further increases it to 91.0\%. The main difference lies in
abnormal-to-normal revisions, which correct 17 false alarms
with bidirectional evidence versus 8 with abnormal evidence
alone, while increasing the number of newly missed anomalies
from 2 to 5. Normal-to-abnormal revisions change little across
the two variants. These results indicate that normal evidence
primarily helps EVAD recover from initial false alarms, with
a modest increase in missed detections.

\section{Conclusion}

In this paper, we presented EVAD, a training-free framework for industrial anomaly detection and reasoning. EVAD integrates semantics-guided inspection planning, bidirectional evidence-guided tree search, and diversity-aware decision reassessment to enable effective acquisition and utilization of informative visual evidence. By combining global visual understanding with active regional exploration, EVAD progressively examines fine-grained visual details through semantic refinement and spatial expansion, while jointly considering normal and abnormal evidence to guide the search. Furthermore, diversity-aware evidence selection reduces redundant observations and facilitates evidence-based refinement of initial predictions. Extensive experiments across multiple industrial anomaly understanding tasks demonstrate the effectiveness of EVAD. Overall, EVAD provides a promising approach to reliable and interpretable industrial anomaly understanding through active visual inspection and evidence-guided reasoning. Future work will explore integrating stronger multimodal backbones to further improve inspection accuracy.

\section*{LLM Usage Statement}

During manuscript preparation, ChatGPT (OpenAI) was used for language polishing and grammatical proofreading. All suggested revisions were reviewed by the authors, who take full responsibility for the final manuscript.

{
    \small
    \bibliographystyle{ieeenat_fullname}
    \bibliography{main}

@String(AAAI = {AAAI})

@inproceedings{bergmann2019mvtec,
  author    = {Bergmann, Paul and Fauser, Michael and Sattlegger, David and Steger, Carsten},
  title     = {{MVTec AD} -- A Comprehensive Real-World Dataset for Unsupervised Anomaly Detection},
  booktitle = {Proceedings of the IEEE/CVF Conference on Computer Vision and Pattern Recognition},
  pages     = {9592--9600},
  year      = {2019}
}

@inproceedings{roth2022towards,
  author    = {Roth, Karsten and Pemula, Latha and Zepeda, Joaquin and
               Sch{\"o}lkopf, Bernhard and Brox, Thomas and Gehler, Peter},
  title     = {Towards Total Recall in Industrial Anomaly Detection},
  booktitle = {Proceedings of the IEEE/CVF Conference on Computer Vision and Pattern Recognition},
  pages     = {14318--14328},
  year      = {2022}
}

@inproceedings{zavrtanik2021draem,
  author    = {Zavrtanik, Vitjan and Kristan, Matej and Sko{\v{c}}aj, Danijel},
  title     = {{DRAEM} -- A Discriminatively Trained Reconstruction Embedding for Surface Anomaly Detection},
  booktitle = {Proceedings of the IEEE/CVF International Conference on Computer Vision},
  pages     = {8330--8339},
  year      = {2021}
}

@inproceedings{jeong2023winclip,
  author    = {Jeong, Jongheon and Zou, Yang and Kim, Taewan and
               Zhang, Dongqing and Ravichandran, Avinash and Dabeer, Onkar},
  title     = {{WinCLIP}: Zero-/Few-Shot Anomaly Classification and Segmentation},
  booktitle = {Proceedings of the IEEE/CVF Conference on Computer Vision and Pattern Recognition},
  pages     = {19606--19616},
  year      = {2023}
}

@inproceedings{zhou2024anomalyclip,
  author    = {Zhou, Qihang and Pang, Guansong and Tian, Yu and
               He, Shibo and Chen, Jiming},
  title     = {{AnomalyCLIP}: Object-Agnostic Prompt Learning for Zero-Shot Anomaly Detection},
  booktitle = {International Conference on Learning Representations},
  year      = {2024}
}

@inproceedings{gu2024anomalygpt,
  author    = {Gu, Zhaopeng and Zhu, Bingke and Zhu, Guibo and
               Chen, Yingying and Tang, Ming and Wang, Jinqiao},
  title     = {{AnomalyGPT}: Detecting Industrial Anomalies Using Large Vision-Language Models},
  booktitle = {Proceedings of the AAAI Conference on Artificial Intelligence},
  volume    = {38},
  number    = {3},
  pages     = {1932--1940},
  year      = {2024},
  doi       = {10.1609/aaai.v38i3.27963}
}

@inproceedings{jiang2025mmad,
  author    = {Jiang, Xi and Li, Jian and Deng, Hanqiu and Liu, Yong and
               Gao, Bin-Bin and Zhou, Yifeng and Li, Jialin and
               Wang, Chengjie and Zheng, Feng},
  title     = {{MMAD}: A Comprehensive Benchmark for Multimodal Large Language Models in Industrial Anomaly Detection},
  booktitle = {International Conference on Learning Representations},
  year      = {2025}
}

@article{bergmann2022beyond,
  author  = {Bergmann, Paul and Batzner, Kilian and Fauser, Michael and
             Sattlegger, David and Steger, Carsten},
  title   = {Beyond Dents and Scratches: Logical Constraints in Unsupervised Anomaly Detection and Localization},
  journal = {International Journal of Computer Vision},
  volume  = {130},
  pages   = {947--969},
  year    = {2022},
  doi     = {10.1007/s11263-022-01578-9}
}

@inproceedings{shen2025zoomeye,
  author    = {Shen, Haozhan and Zhao, Kangjia and Zhao, Tiancheng and
               Xu, Ruochen and Zhang, Zilun and Zhu, Mingwei and Yin, Jianwei},
  title     = {{ZoomEye}: Enhancing Multimodal {LLMs} with Human-Like Zooming Capabilities through Tree-Based Image Exploration},
  booktitle = {Proceedings of the 2025 Conference on Empirical Methods in Natural Language Processing},
  pages     = {6602--6618},
  year      = {2025},
  doi       = {10.18653/v1/2025.emnlp-main.335}
}

@article{peng2026eagle,
  author  = {Peng, Xiaomeng and Huang, Xilang and Choi, Seon Han},
  title   = {{EAGLE}: Expert-Augmented Attention Guidance for Tuning-Free Industrial Anomaly Detection in Multimodal Large Language Models},
  journal = {arXiv preprint arXiv:2602.17419},
  year    = {2026}
}

@inproceedings{carion2026sam3,
  author    = {Carion, Nicolas and Gustafson, Laura and Hu, Yuan-Ting and
               Debnath, Shoubhik and Hu, Ronghang and Suris Coll-Vinent, Didac and
               Ryali, Chaitanya and Alwala, Kalyan Vasudev and Khedr, Haitham and
               Huang, Andrew and Lei, Jie and Ma, Tengyu and Guo, Baishan and
               Kalla, Arpit and Marks, Markus and Greer, Joseph and Wang, Meng and
               Sun, Peize and R{\"a}dle, Roman and Afouras, Triantafyllos and
               Mavroudi, Effrosyni and Xu, Katherine and Wu, Tsung-Han and
               Zhou, Yu and Momeni, Liliane and Hazra, Rishi and Ding, Shuangrui and
               Vaze, Sagar and Porcher, Francois and Li, Feng and Li, Siyuan and
               Kamath, Aishwarya and Cheng, Ho Kei and Doll{\'a}r, Piotr and
               Ravi, Nikhila and Saenko, Kate and Zhang, Pengchuan and
               Feichtenhofer, Christoph},
  title     = {{SAM 3}: Segment Anything with Concepts},
  booktitle = {International Conference on Learning Representations},
  year      = {2026}
}

@inproceedings{kocsis2006bandit,
  author    = {Kocsis, Levente and Szepesv{\'a}ri, Csaba},
  title     = {Bandit Based Monte-Carlo Planning},
  booktitle = {European Conference on Machine Learning},
  series    = {Lecture Notes in Computer Science},
  volume    = {4212},
  pages     = {282--293},
  publisher = {Springer},
  year      = {2006},
  doi       = {10.1007/11871842_29}
}

@article{bai2025qwen3vl,
  author  = {Bai, Shuai and Cai, Yuxuan and Chen, Ruizhe and Chen, Keqin and
             Chen, Xionghui and Cheng, Zesen and Deng, Lianghao and Ding, Wei and
             Gao, Chang and Ge, Chunjiang and Ge, Wenbin and Guo, Zhifang and
             Huang, Qidong and Huang, Jie and Huang, Fei and Hui, Binyuan and
             Jiang, Shutong and Li, Zhaohai and Li, Mingsheng and Li, Mei and
             Li, Kaixin and Lin, Zicheng and Lin, Junyang and Liu, Xuejing and
             Liu, Jiawei and Liu, Chenglong and Liu, Yang and Liu, Dayiheng and
             Liu, Shixuan and Lu, Dunjie and Luo, Ruilin and Lv, Chenxu and
             Men, Rui and Meng, Lingchen and Ren, Xuancheng and Ren, Xingzhang and
             Song, Sibo and Sun, Yuchong and Tang, Jun and Tu, Jianhong and
             Wan, Jianqiang and Wang, Peng and Wang, Pengfei and Wang, Qiuyue and
             Wang, Yuxuan and Xie, Tianbao and Xu, Yiheng and Xu, Haiyang and
             Xu, Jin and Yang, Zhibo and Yang, Mingkun and Yang, Jianxin and
             Yang, An and Yu, Bowen and Zhang, Fei and Zhang, Hang and Zhang, Xi and
             Zheng, Bo and Zhong, Humen and Zhou, Jingren and Zhou, Fan and
             Zhou, Jing and Zhu, Yuanzhi and Zhu, Ke},
  title   = {{Qwen3-VL} Technical Report},
  journal = {arXiv preprint arXiv:2511.21631},
  year    = {2025}
}

@inproceedings{li2025triad,
  author    = {Li, Yuanze and Yuan, Shihao and Wang, Haolin and Li, Qizhang and
               Liu, Ming and Xu, Chen and Shi, Guangming and Zuo, Wangmeng},
  title     = {{Triad}: Empowering {LMM}-Based Anomaly Detection with
               Expert-Guided Region-of-Interest Tokenizer and Manufacturing Process},
  booktitle = {Proceedings of the IEEE/CVF International Conference on Computer Vision},
  pages     = {21917--21926},
  year      = {2025}
}

@inproceedings{li2026iadr1,
  author    = {Li, Yanhui and Cao, Yunkang and Liu, Chengliang and
               Xiong, Yuan and Dong, Xinghui and Huang, Chao},
  title     = {{IAD-R1}: Reinforcing Consistent Reasoning in Industrial Anomaly Detection},
  booktitle = {Proceedings of the AAAI Conference on Artificial Intelligence},
  volume    = {40},
  number    = {8},
  pages     = {6583--6591},
  year      = {2026},
  doi       = {10.1609/aaai.v40i8.37588}
}

@inproceedings{liao2026adfm,
  author    = {Liao, Jingyi and Su, Yongyi and Tu, Rong-Cheng and Jin, Zhao and
               Sun, Wenhao and Li, Yiting and Tao, Dacheng and Xu, Xun and Yang, Xulei},
  title     = {{AD-FM}: Multimodal {LLMs} for Anomaly Detection via Multi-Stage
               Reasoning and Fine-Grained Reward Optimization},
  booktitle = {Proceedings of the AAAI Conference on Artificial Intelligence},
  volume    = {40},
  number    = {18},
  pages     = {15234--15242},
  year      = {2026},
  doi       = {10.1609/aaai.v40i18.38548}
}

@inproceedings{zhang2026adseeker,
  author    = {Zhang, Kai and Zhang, Zekai and Sun, Xihe and Wang, Anpeng and
               Nie, Jingmeng and Chen, Qinghui and Hao, Han and
               Guo, Jianyuan and Zhang, Jinglin},
  title     = {{ADSeeker}: A Knowledge-Grounded Reasoning Framework for
               Industry Anomaly Detection and Reasoning},
  booktitle = {Proceedings of the IEEE/CVF Conference on Computer Vision and Pattern Recognition},
  pages     = {21379--21388},
  year      = {2026}
}

@inproceedings{wu2024vstar,
  author    = {Wu, Penghao and Xie, Saining},
  title     = {{V*}: Guided Visual Search as a Core Mechanism in Multimodal {LLMs}},
  booktitle = {Proceedings of the IEEE/CVF Conference on Computer Vision and Pattern Recognition},
  pages     = {13084--13094},
  year      = {2024}
}

@inproceedings{hu2024visual,
  author    = {Hu, Yushi and Shi, Weijia and Fu, Xingyu and Roth, Dan and
               Ostendorf, Mari and Zettlemoyer, Luke and Smith, Noah A. and
               Krishna, Ranjay},
  title     = {{Visual Sketchpad}: Sketching as a Visual Chain of Thought for
               Multimodal Language Models},
  booktitle = {Advances in Neural Information Processing Systems},
  volume    = {37},
  pages     = {139348--139379},
  year      = {2024}
}

@inproceedings{wang2025visuothink,
  author    = {Wang, Yikun and Wang, Siyin and Cheng, Qinyuan and Fei, Zhaoye and
               Ding, Liang and Guo, Qipeng and Tao, Dacheng and Qiu, Xipeng},
  title     = {{VisuoThink}: Empowering {LVLM} Reasoning with Multimodal Tree Search},
  booktitle = {Proceedings of the 63rd Annual Meeting of the Association for
               Computational Linguistics (Volume 1: Long Papers)},
  pages     = {21707--21719},
  year      = {2025},
  doi       = {10.18653/v1/2025.acl-long.1053}
}

@misc{anthropic2024claude35sonnet,
  author = {{Anthropic}},
  title  = {{Claude 3.5 Sonnet} Model Card Addendum},
  year   = {2024},
  url    = {https://assets.claude.com/fed9cc193a14b84131812372d8d5857f8f304c52.pdf}
}

@article{comanici2025gemini25,
  author  = {Comanici, Gheorghe and Bieber, Eric and Schaekermann, Mike and
             Pasupat, Ice and Sachdeva, Noveen and others},
  title   = {{Gemini 2.5}: Pushing the Frontier with Advanced Reasoning,
             Multimodality, Long Context, and Next Generation Agentic Capabilities},
  journal = {arXiv preprint arXiv:2507.06261},
  year    = {2025}
}

@misc{openai2024gpt4o,
  author = {{OpenAI}},
  title  = {{GPT-4o} System Card},
  year   = {2024},
  url    = {https://openai.com/index/gpt-4o-system-card/}
}

@misc{openai2026gpt55,
  author = {{OpenAI}},
  title  = {{GPT-5.5} System Card},
  year   = {2026},
  url    = {https://openai.com/index/gpt-5-5-system-card/}
}

@article{li2025llavaonevision,
  author  = {Li, Bo and Zhang, Yuanhan and Guo, Dong and Zhang, Renrui and
             Li, Feng and Zhang, Hao and Zhang, Kaichen and Zhang, Peiyuan and
             Li, Yanwei and Liu, Ziwei and Li, Chunyuan},
  title   = {{LLaVA-OneVision}: Easy Visual Task Transfer},
  journal = {Transactions on Machine Learning Research},
  year    = {2025}
}

@article{zhu2025internvl3,
  author  = {Zhu, Jinguo and Wang, Weiyun and Chen, Zhe and Liu, Zhaoyang and
             Ye, Shenglong and others},
  title   = {{InternVL3}: Exploring Advanced Training and Test-Time Recipes
             for Open-Source Multimodal Models},
  journal = {arXiv preprint arXiv:2504.10479},
  year    = {2025}
}

@article{glmteam2025glmv,
  author  = {{GLM-V Team}},
  title   = {{GLM-4.5V} and {GLM-4.1V-Thinking}: Towards Versatile Multimodal
             Reasoning with Scalable Reinforcement Learning},
  journal = {arXiv preprint arXiv:2507.01006},
  year    = {2025}
}

@article{kimiteam2025kimivl,
  author  = {{Kimi Team}},
  title   = {{Kimi-VL} Technical Report},
  journal = {arXiv preprint arXiv:2504.07491},
  year    = {2025}
}

@inproceedings{zhang2025eiad,
  author    = {Zhang, Zongyun and Ruan, Jiacheng and Gao, Xian and
               Liu, Ting and Fu, Yuzhuo},
  title     = {{EIAD}: Explainable Industrial Anomaly Detection via
               Multi-Modal Large Language Models},
  booktitle = {IEEE International Conference on Multimedia and Expo},
  pages     = {1--6},
  year      = {2025},
  doi       = {10.1109/ICME59968.2025.11209158}
}

@article{chao2025anomalyr1,
  author  = {Chao, Yuhao and Liu, Jie and Tang, Jie and Wu, Gangshan},
  title   = {{AnomalyR1}: A {GRPO}-Based End-to-End {MLLM} for
             Industrial Anomaly Detection},
  journal = {arXiv preprint arXiv:2504.11914},
  year    = {2025}
}

@inproceedings{zou2022spot,
  author    = {Zou, Yang and Jeong, Jongheon and Pemula, Latha and
               Zhang, Dongqing and Dabeer, Onkar},
  title     = {{SPot-the-Difference} Self-Supervised Pre-Training for
               Anomaly Detection and Segmentation},
  booktitle = {European Conference on Computer Vision},
  pages     = {392--408},
  year      = {2022},
  doi       = {10.1007/978-3-031-20056-4_23}
}

@article{zhang2024goodsad,
  author  = {Zhang, Jian and Ding, Runwei and Ban, Miaoju and Yang, Ge},
  title   = {{PKU-GoodsAD}: A Supermarket Goods Dataset for Unsupervised
             Anomaly Detection and Segmentation},
  journal = {IEEE Robotics and Automation Letters},
  volume  = {9},
  number  = {3},
  pages   = {2008--2015},
  year    = {2024},
  doi     = {10.1109/LRA.2024.3352358}
}

@article{chen2026dyc,
  title={DyC-CLIP: Dynamic context-aware multi-modal prompt learning for zero-shot anomaly detection},
  author={Chen, Peng and Huang, Fangjun and Huang, Chao},
  journal={Pattern Recognition},
  pages={113215},
  year={2026},
  publisher={Elsevier}
}

@article{li2026bvs,
  title={BVS: Bayesian Visual Search with Multimodal Large Language Model for Fine-grained Perception},
  author={Li, Geng and Peng, Yuxin},
  journal={arXiv preprint arXiv:2607.03184},
  year={2026}
}

@inproceedings{gao2026adaptclip,
  title={Adaptclip: Adapting clip for universal visual anomaly detection},
  author={Gao, Bin-Bin and Zhou, Yue and Yan, Jiangtao and Cai, Yuezhi and Zhang, Weixi and Wang, Meng and Liu, Jun and Liu, Yong and Wang, Lei and Wang, Chengjie},
  booktitle={Proceedings of the AAAI Conference on Artificial Intelligence},
  volume={40},
  number={6},
  pages={4095--4103},
  year={2026}
}

@inproceedings{cao2024adaclip,
  title={Adaclip: Adapting clip with hybrid learnable prompts for zero-shot anomaly detection},
  author={Cao, Yunkang and Zhang, Jiangning and Frittoli, Luca and Cheng, Yuqi and Shen, Weiming and Boracchi, Giacomo},
  booktitle={European conference on computer vision},
  pages={55--72},
  year={2024},
  organization={Springer}
}

@article{chen2026towards,
  title={Towards Explainable Industrial Anomaly Detection via Knowledge-Guided Latent Reasoning},
  author={Chen, Peng and Huang, Chao and Cao, Yunkang and Liu, Chengliang and Wang, Wei and Wang, Wenqiang and Yang, Mingbo and Shen, Li and Ren, Wenqi and Cao, Xiaochun},
  journal={arXiv preprint arXiv:2602.09850},
  year={2026}
}

@article{xiaomi2025mimovl,
  author  = {{Xiaomi LLM-Core Team}},
  title   = {{MiMo-VL} Technical Report},
  journal = {arXiv preprint arXiv:2506.03569},
  year    = {2025}
}

@inproceedings{li2026deepscan,
  title={Deepscan: A training-free framework for visually grounded reasoning in large vision-language models},
  author={Li, Yangfu and Zhan, Hongjian and Chen, Jiawei and Gong, Yuning and Liu, Qi and Lu, Yue},
  booktitle={Proceedings of the IEEE/CVF Conference on Computer Vision and Pattern Recognition},
  pages={19155--19164},
  year={2026}
}

@article{liu2025hide,
  title={Hide: Rethinking the zoom-in method in high resolution mllms via hierarchical decoupling},
  author={Liu, Xianjie and Hu, Yiman and Zou, Yixiong and Wu, Liang and Xu, Jian and Zheng, Bo},
  journal={arXiv preprint arXiv:2510.00054},
  year={2025}
}
}


\end{document}